# "Training Convolutional Networks with Web Images"

**Masters Thesis by**

## Nizar Massouh

**Under the supervision of**

## Barbara Caputo

**7/1/2016**

Training  Convolutional  Networks  with  Web  Images  −  Nizar  Massouh

# Table of Contents

Training  Convolutional  Networks  with  Web  Images  –  Nizar  Massouh

# Chapter 1

# Introduction

In the past decade Computer Vision has evolved enormously with the availability of cameras and the rise of demand on visual applications. The spread of the computer vision applications to other connected disciplines (Artificial Intelligence, Machine Learning, Robotics, Image Processing, Neural Network, Computer graphics... etc.) stimulated the growth of available technology and opened the door for new discoveries. The increased use of machine learning and the evolution of the processing power and memory was very beneficiary for Computer Vision. One of the most important sub-domains of Computer Vision is Object recognition that is a learning-based method that relies on images. Trying to extract useful information from images has proved to be a complex and challenging task that has employed the computer vision community for decades [30]. Learning from images goes through a series of image related challenges that include: *the camera position*, where changing the position of the camera can alter the viewpoint that can make it harder to recognize certain objects. *Illumination* plays a very important role in images since the presence or absence of it can transform the whole scene or the pixel intensity. *Occlusion* and *background clutter* can occur in scenes where a part of an object is partially hidden or surround by a lot noise which can make the task of extracting useful information challenging. Another tough situation is *Intra-class variation* where the design or features of an object can change a great deal from one subject to another (figure 1).

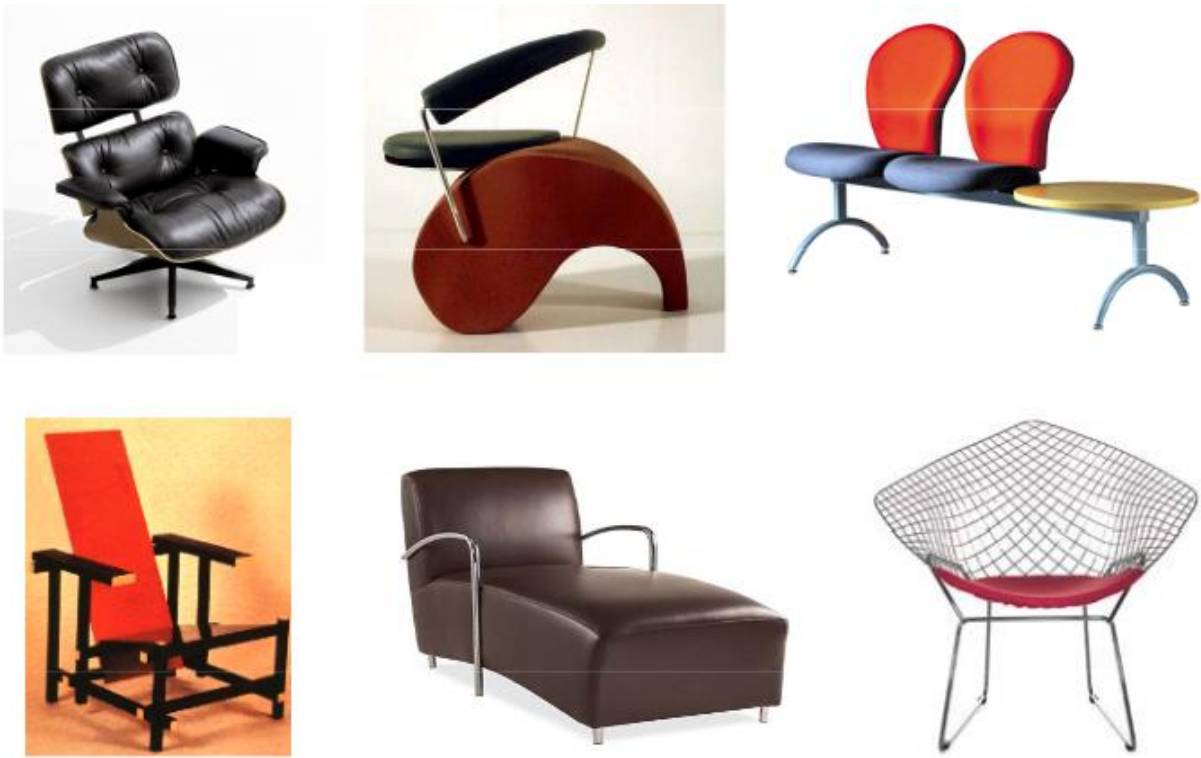

**Figure 1 shows an example of Intra-Class variation on the Chair class.**

*But how will a machine learn from images?* Visual learning consists of trying to learn what distinguish one image from another by extracting the features of each image. Trying to grasp a concept from images requires a lot of training data that serve as a reference. For example if our machine is trying to learn to recognize dogs, it will need to see a lot of examples of dogs through different images that form a training set to be able to create a model of that object (dog). There exist many techniques to actually "learn" from images, using the extracted features on a set of algorithms we want to be able to recognize an object or classify it and output our result. Learning methods can vary from simple "shallow" to complex and "deep" methods. Shallow learning methods take the extracted feature vector of an image as input and combine them with weights based on all the training data seen can draw a conclusion and categorize the fed image. This technique relies on external tools to process the images and extract their features to be used. The simplicity of such methods (shallow) tends to have some limitations when confronted with multiple concepts; this is where deep learning comes in handy. Deep learning [31] unlike other approaches will take in the raw images as input and will learn the features to a very high level of abstraction that will be used to discriminate against new images for recognition tasks. Deep learning transforms its input through a series of layered processing units which learns multiple levels of representations which deciphers the image and is therefore able to learn it on a deeper level than that of Shallow learners. This method has been very successful for Visual Recognition tasks but it comes at a price since it needs long computational time and Large Scale data to be able to generalize well. Large Scale databases are built with millions of images that are usually labelled and annotated using human experts. Having human annotators to work on millions of images to build a database makes the process costly. In this thesis we hope to investigate the effect of using web images to build a large scale database to be used along a deep learning method for a classification task. Therefore we chose to replicate the ImageNet large scale database (`ILSVRC-2012`) [4] from images downloaded from the web using 4 different collection strategies varying: the search engine, the query and the image resolution. As a deep learning method, we will choose the Convolutional Neural Network that was very successful with recognition tasks; the AlexNet [11]. Using as Benchmark the results of [21], we reproduce the two experiments done to test our data on Object recognition and Domain adaptation tasks. Our collected data provide a big assortment of representations that while they could not provide great results as training data [32] [33], has proven to produce a feature extractor that was able to abstract better than the extractor trained on ImageNet.

The rest of the thesis is structured as follows: *Chapter 2* reviews previous work in the literature, placing our work in context of current research in the field. *Chapter 3* describes the data download process, shedding the light on the four collected databases. *Chapter 4* explains the deep learning method and a comprehensive look into AlexNet's configuration. *Chapter 5* will take us through the experimental procedures and results. In *chapter 6* we will conclude our thesis with some analysis and explains what more can be done.

# Chapter 2

# Related Work

To successfully learn about an object using millions of images requires a model that can handle the magnitude of the task. Convolutional Neural Networks (CNNs) [34] have what it takes to handle the complexity of the object recognition task since they can be built and tuned to accommodate the millions parameters required. CNNs have realized the state of the art performance for Image classification tasks on Large Scale data [11] but the learned feature extractor can be applied to other jobs as well. Transfer learning has been studied using deep networks using unsupervised and supervised settings [35]. In Transfer learning a CNN is pre-trained on a Large Scale database and transferring the learned feature extractor to different tasks with different classes/labels. Another aspect of the CNN's learned features that has been evaluated and studied is how well they generalize. To evaluate such characteristic the features are applied to a Domain Adaptation experiment [27] that tests it on different domain switches.

A CNN requires Large Scale data to successfully learn the variability and abstract representation of objects, therefore having large scale labeled databases that contain millions of images with a variety of classes is required. But other research has been conducted on different types of data like synthetic images [36] or web images [37]. Synthetic images have been used for many computer vision tasks but current trends researchers have used 3D computer generated images [38] to detection and recognition tasks. Another trend is to use Web data to train CNNs or to learn visual representation [39]. Our work is inspired by all the above and we hope throughout this thesis to contribute in the discovery of the full potentials of Web images and their effect on CNNs.

# Chapter 3

# Data download

Our goal is to provide the necessary tools to build large scale databases from images available on the public domain. We would like to have an automatic system that can construct databases without the help of human annotators while staying relevant to the topics they include. One of the biggest and mostly used large scale databases in visual recognition is Imagenet [6]. Imagenet is an image database that currently contains up to 14 million images structured in almost 21 thousands categories (figure 2). All images provided by Imagenet are quality controlled and human-annotated. To push the community, the Imagenet Large Scale Visual Recognition challenges (ILSVRC) has been organized since 2010 [4]. The challenge evaluates results on object detection tasks and image classification for large scale. Since the beginning the challenges have been implemented on a 1000 object categories' database as training data, and 50000 labeled images used as a validation set. We would like to be able to get enough images, without the help of human annotators, per class to be larger or equal to those of ImageNet's ILSVRC-2012 1000 objects classes' database, which currently contains around 1.2 million images [2].

ImageNet uses the WordNet [5] hierarchy as structure to their data. Each concept in WordNet has either one or a list of words that describes it; they are called Synset (Synonym set). ImageNet provides almost 1000 images per concept/class on average. Currently WordNet contains around 100,000 concepts (mostly nouns). Understanding the structure of WordNet will help us on our task to create an efficient Query list to acquire the images online. The 1000 object categories' database is formed by mutually exclusive leaf nodes of the WordNet hierarchy tree.

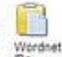

**Figure 2 imagenet 'Pedestrian crossing' category example**

## 3.1 Query Structure

ImageNet's 1000 objects database contains, on average, a total of 1281 images per class. The most popular image search engines available (google, yahoo, bing etc.) can provide up to 1000 images per query with the assumption that the number will suffice for non-commercial users, which is not enough alone to recreate Imagenet. Considering that duplicated results might occur and some categories could return a

Training Convolutional Networks with Web Images − Nizar Massouh

very low number of images, a query augmentation strategy has to be implemented. Since our focus is not about the creation of a query expansion algorithm that can beat the state of the art, we have devised a plan to expand the queries in a very simple way. Our method relies on the WordNet hierarchy tree.

All words in WordNet are essentially connected by synonyms which form sets (Synsets). There are currently around 117,000 synsets which have certain relations between them. The relation that is mostly useful for is called 'hypernym' or an IS_A relation.

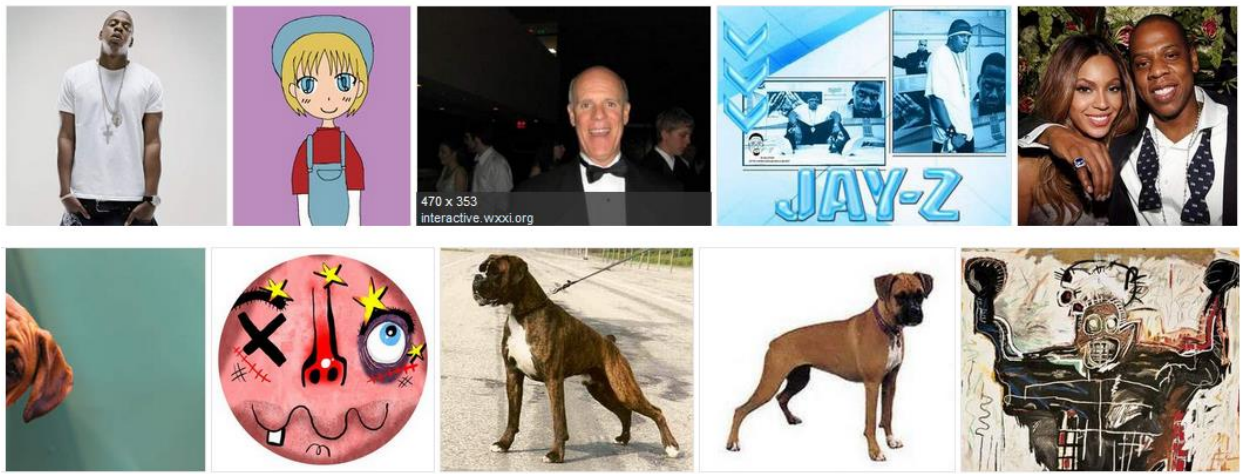

Figure 3. Top results of misleading query on Picsearch: Top is the result of the query 'Jay', Bottom is  the result of the query 'boxer'

Our method comprises of 2 steps: the first is taking the synset of each class and expanding it by appending the parent of the node to it. The whole point is to try to gain more images while keeping them as relevant as possible. The relevance of the query varies widely from one category to another, some are very easy (pizza) and some are very difficult (boxer or Jay). As we can see in figure 3, two different kind of misleading queries. 'Jay' as a query is misleading because of its incompleteness; since it is a proper noun it cannot represent one subject if used alone. Whereas 'Boxer' as a noun has multiple synonyms (dog, martial art fighter…) which can result in a word-sense disambiguation problem [7] if not coupled with the right keywords that can filter out unwanted results. Let's take a look at how our method can help:

The category 'Jay' has no synonyms, therefore no synset. Using the query 'Jay' on a search engine and expecting images of birds is a little bit far-fetched. The query alone can be very misleading, but if we add 'bird' to it we will be filtering the results to much more relevant data. As we can see in the Jay example (figure 4), the word "bird" exists as a parent. The same thing applies to other queries. The branch that leads to Jay is as follow:

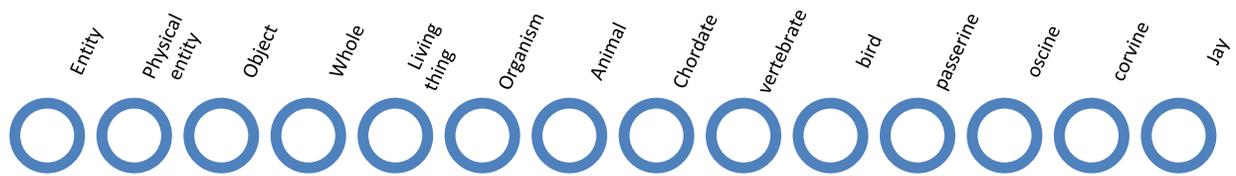

**Figure 4. Branch off the tree of the leaf 'Jay'**

Other examples of misleading data exist（Boxer,Crane,Crane…etc）and they can all be fixed by appending the parent to the query（'Boxer dog', 'Crane bird', 'Crane lift'…etc）.

Adding the parent alone is sufficient to make the query relevant, but we need to be able to augment it as well. Let's go back to our Jay example. 'Jay' has no synonyms, therefore 'Jay bird' alone wouldn't give us the minimum we need for the experiment. That's where the second step of our method comes in handy. Our second step is to translate the query to other languages（with the parent already appended）. After having downloaded using the query of step one, in the cases where the images are not sufficient we use step two. The astonishing fact is for example 'Jay Bird' results with 700 images from Google; whereas its Spanish equivalent, 'Jay pájaro', gets a totally new set of data（figure 5）. Google image result is returned if the query we are searching for is present in the text associated to the image（found on

Training Convolutional Networks with Web Images – Nizar Massouh

websites) or in the image filename or in added image information in the sitemap [8]. This means google does not translate our query to fit certain image information nor the opposite which we will use to our advantage. Since our task doesn't focus entirely on the quantity of the data but the quality, we have used the second step only to enlarge lacking categories.

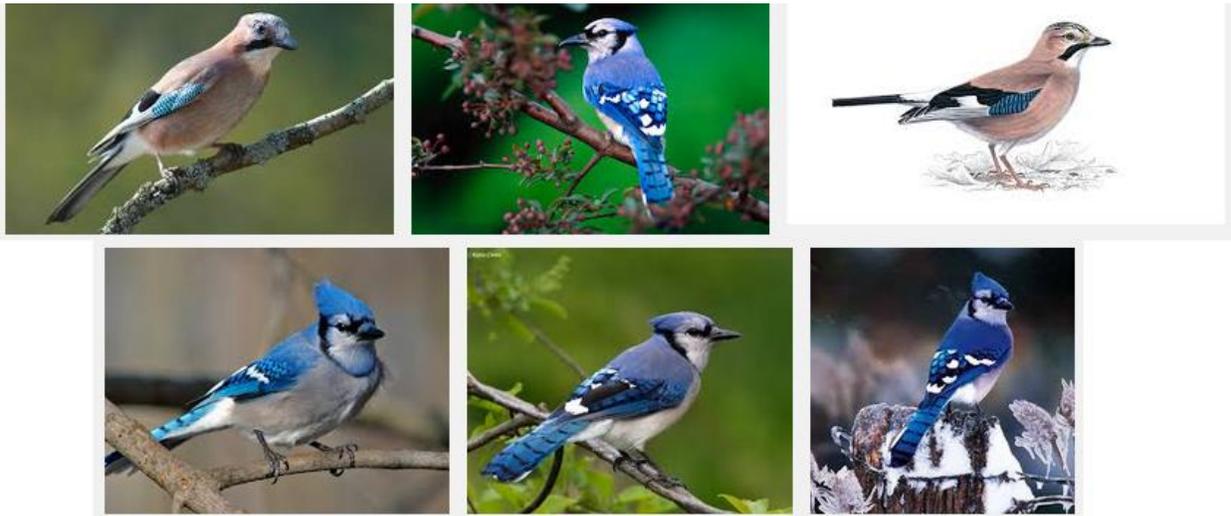

**Figure 5. Top 3 results from Google using:** *Top* **is the result with 'Jay bird',** *Bottom* **is the result with 'Jay pajaro'**

## 3.2 The download process

Ten lists of queries were created to be used for the download. Each list contains 100 queries and some queries contain more than one keyword separated by a comma

Training Convolutional Networks with Web Images – Nizar Massouh

（i.e.  Grey  whale,  gray  whale,  devilfish,  Eschrichtius  gibbosus,  Eschrichtius  robustus）.

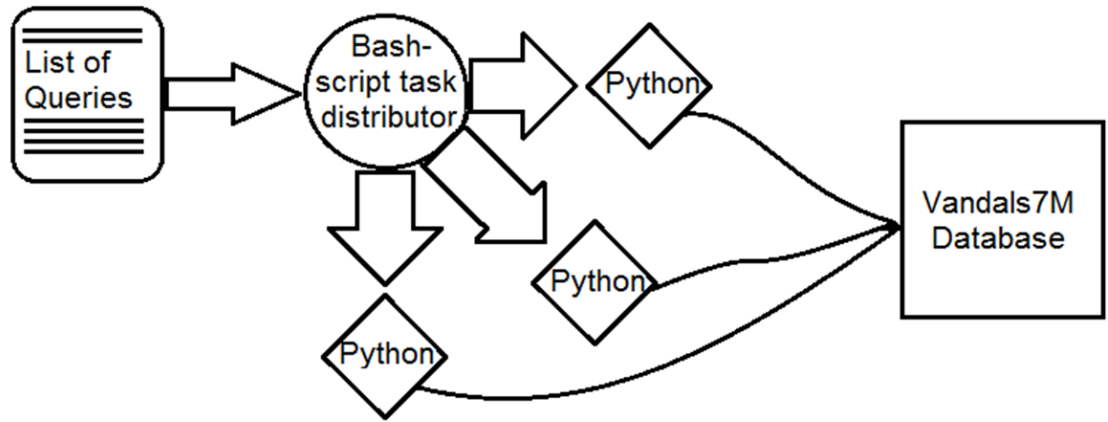

**Figure 6. Images Download Scheme**

A  Bash  script  that  is  fed  a  list  of  queries  will  start  executing  a  python  script  per  line
（figure  6）.  If  a  multiple  keyword  was  detected  then  each  will  be  launched  separately
and  will  be  downloaded  to  the  same  directory.  The  python  script  takes  a  query  as  an
argument  and  collects  the  entire  images  URLs  that  are  returned  from  the  search.
Once  there's  no  more  search  pages  available  or  if  the  list  is  already  larger  than
10000,  a  directory  with  the  class  name  will  be  created  and  the  download  of  the  list
will  start.  We  chose  to  launch  4  lists  at  same  time  to  accelerate  the  process  without
overloading  the  machine.

Training  Convolutional  Networks  with  Web  Images  −  Nizar  Massouh

## 3.3 Data Cleaning

Since some classes have multiple query names that would result in duplicates，as a cleaning step we decided to check all the classes for exact duplicates and for very similar images．For this task we used Perceptual Hashing （ImageHash）[1]．Cryptographic Hash is like a unique ID or fingerprint that defines a certain file；it is used to find duplicates or to differentiate between identical files．

1. **Reduce size**. The fastest way to remove high frequencies and detail is to shrink the image. In this case, shrink it to 8x8 so that there are 64 total pixels. Don't bother keeping the aspect ratio, just crush it down to fit an 8x8 square. This way, the hash will match any variation of the image, regardless of scale or aspect ratio.

   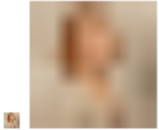

2. **Reduce color**. The tiny 8x8 picture is converted to a grayscale. This changes the hash from 64 pixels (64 red, 64 green, and 64 blue) to 64 total colors.
3. **Average the colors**. Compute the mean value of the 64 colors.
4. **Compute the bits**. This is the fun part. Each bit is simply set based on whether the color value is above or below the mean.
5. **Construct the hash**. Set the 64 bits into a 64-bit integer. The order does not matter, just as long as you are consistent. (I set the bits from left to right, top to bottom using big-endian.)

   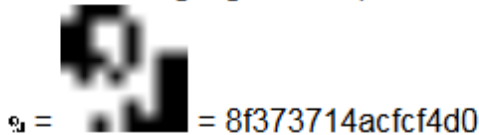 = 8f373714acfcf4d0

Figure 7 Perceptual Hash steps [3]

Cryptographic Hash for images works at the pixel level and it would give a string of characters to identify the image．If two images are "visually" exact but have few pixels that are colored differently this would result in a completely different hash. Therefore we used perceptual hashing or ImageHash which takes the 2 images and decrease their scale (figure 7). By lowering the scaling we switch our focus from small details in more pixels to the global representation on fewer pixels. ImageHash will secure us that no new hash ids will be given to images with very little differences.

## 3.4 Examples [1]:

In this section we illustrate how Image hash works on a 2 different examples.
**1.**

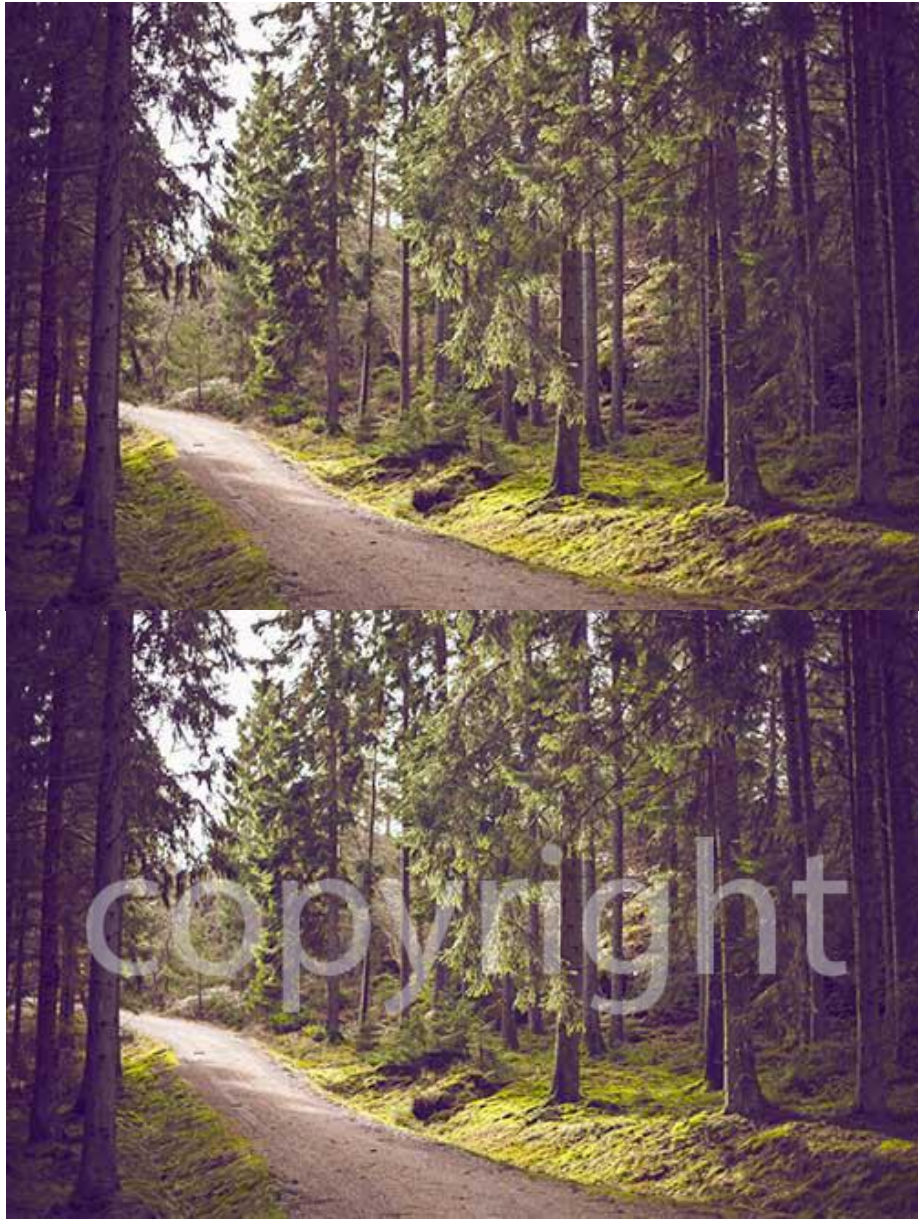

Image 1 hash: **3c3e0e1a3a1e1e1e** (0011110000111110000011100011010001110100001111000011110000011110)
Image 2 hash: **3c3e0e3e3e1e1e1e** (0011110000111110000011100011110001111100001111000011110000011110)

Training Convolutional Networks with Web Images − Nizar Massouh

As seen above these 2 images are quite similar and that can be figured out by looking at their perceptual hashes.

**2.**

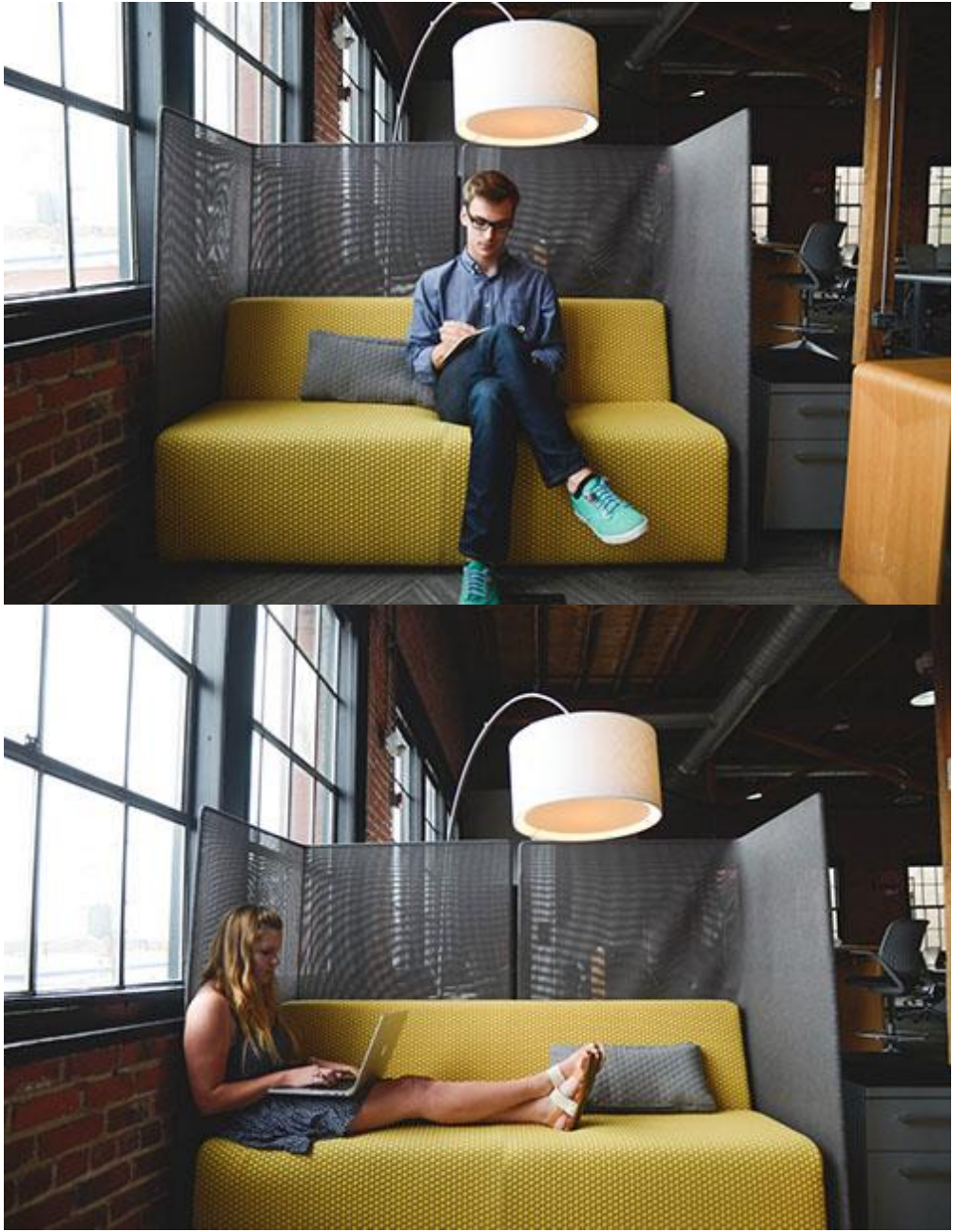

Training Convolutional Networks with Web Images – Nizar Massouh

```
Image 1 hash: 69684858535b7575 (0010100010101000101010001010100010101011001010110101010111001101111)
Image 2 hash: e1e1e2a7bbaf6faf (0111000011110000111100101101001101011011011010010011010101001111)
Hamming distance: 32
```

On the other hand these 2 images aren't similar, thus the large distance.

## 3.5 Collected Data

| DB Name | Query structure | Search engine | Split type | Average number of images per class | Minimum number of images per class | Maximum number of images per class | total number | Notes |
|---|---|---|---|---|---|---|---|---|
| **Random Split** | Synsets | Picsearch | Random from all data | 7326 | 1321 | 42168 | 7.3 million | Focus on Quantity |
| **Controlled Split** | Synsets | Picsearch | Controlled chronological | 7326 | 1321 | 42168 | 7.3 million | Focus on Quality from Quantity |
| **Controlled Download** | Synsets + Parents（unsupervised） | Google+ yahoo | Random from all data | 2494.49 | 1265 | 7919 | 2.4 million | Focus on Quality from search result |
| **Controlled Query** | Synsets + parents（semi-supervised） | Google+ Yahoo+ Flickr | Random from all data | 1911.28 | 923 | 5658 | 1.9 million | Focus on Quality from search and image quality（High Resolution） |

**Table 1 our collected databases' description**

As we can see in table 1, each database presents a very different outlook on the data and holds promise of showing diverse results. We have started out with Picsearch, and we downloaded smaller versions of the images while trying to get as many as possible. This approach gave a very big amount of data（7.3 million images）using only the synsets as query. The choice to commit to only the synsets,

Training Convolutional Networks with Web Images – Nizar Massouh

knowing that they contain many misleading ones, was to try to test the method on a totally automatic process. The Controlled Split, unlike the Random split, follows the assumption that the relevance of the search results decreases the further we advance with the results. With that in mind, we chose our images following the chronological order of which they were downloaded. The Controlled Download split applied an unsupervised query expansion method that appends the name of the parent category to the synsets. Moreover, it does not request as many images as the search returns; but it stops after a certain number. Our last dataset, Controlled Query, as the name indicates it has a semi-supervised query expansion. After using the automatic method of adding parents to the synsets, we manually controlled the parents and made them relevant to the category trying to make them as visual as possible. For the last dataset we decided to download the images in high resolution (figure 8).

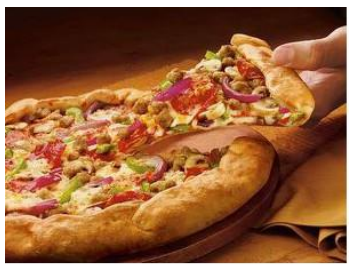 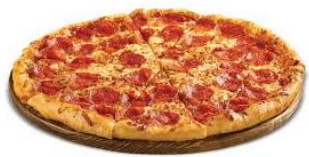 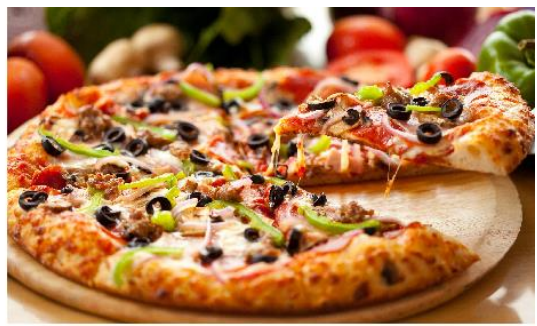

299x244        316x159        2560x1600

**Figure 8. Different examples of the category Pizza from: Left, RandomSplit db. Center, ControlledDownload db. Right, ControlledQuery db.**

Training Convolutional Networks with Web Images – Nizar Massouh

# Chapter 4

# Classification using Convolutional Neural Network

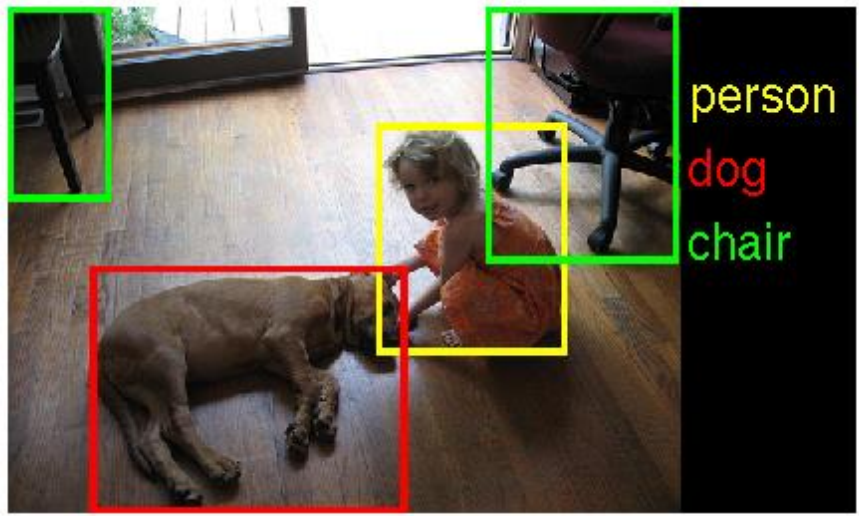

**Figure 9 An Object classification example**

## 4.1 Object Classification from Images

One of the main goals of machine learning is to develop the technology to allow robots or AI agents to learn complex concepts with low or no human supervision. Identifying objects in their environment is an example of a complex task of interest to us. Object classification from images is one of the most challenging tasks in computer vision. A lot of progress in the past years has helped developing methods to extract the characteristics of images, features, and then apply machine learning techniques to those features. To be able to classify an object we need to have enough knowledge of that object to successfully detect and recognize it (figure 9).

Therefore a set of data should be used for training the classifier. Images used for training should include many examples of the object to help form a robust representation of that object. All the images need to be transformed into a representation that maximizes the task at hands: extracting the most relevant features of the data, enhancing them and discarding the data that can be noisy or out of topic. Many feature extractors have been developed to translate the information in images into object oriented discriminative data (sift, HOG...etc.). Using a classifier that can take as input the extracted features and using a set of algorithms can derive an output: the object's class. The better the features are represented the easier the classification becomes. While we strive to make the classifier learn a certain concept completely unguided by humans or what is called an ***unsupervised learning***, there exist other learning methods that include some help. A learning strategy is called ***fully supervised*** when data passed to a classifier is fully labeled therefore the classifier can easily associate the features extracted to the class indicated by the label. A *semi-supervised* method means that the classifier will have some labeled examples in its training set which will also help it to better classify similar features to those known examples. A supervised learning algorithm [17] needs a training set of $N$ examples of the form $\{(x_1, y_1) \dots (x_N, y_N)\}$ where $x_i$ is the i-th image and $y_i$ is its true label or class ($i=1 \dots$ N and $y_i = 1 \dots$ C) and then using a scoring function $f$ that returns the output $y$ that gives the highest score, this algorithm is best represented as:

$$g(x) = \arg\max_y \quad f(x, y) \qquad (1)$$

In the formula (1) $g$ is an element of the hypothesis space, which means the space where all the possible functions that can best approximate the target function $f$ or in other words the function that if given an input $x$ will return the output $y$ such that:

$$y = f(x) \qquad (2)$$

The task of object classification remains to this day a big stimulant to all scientists that have made the state of the art as close to the human results as much as the current technology allows [9]. There exist many types of classifiers that can be used to classify objects from images. The ones that interest us are shallow classifiers and deep classifiers (figure 10).

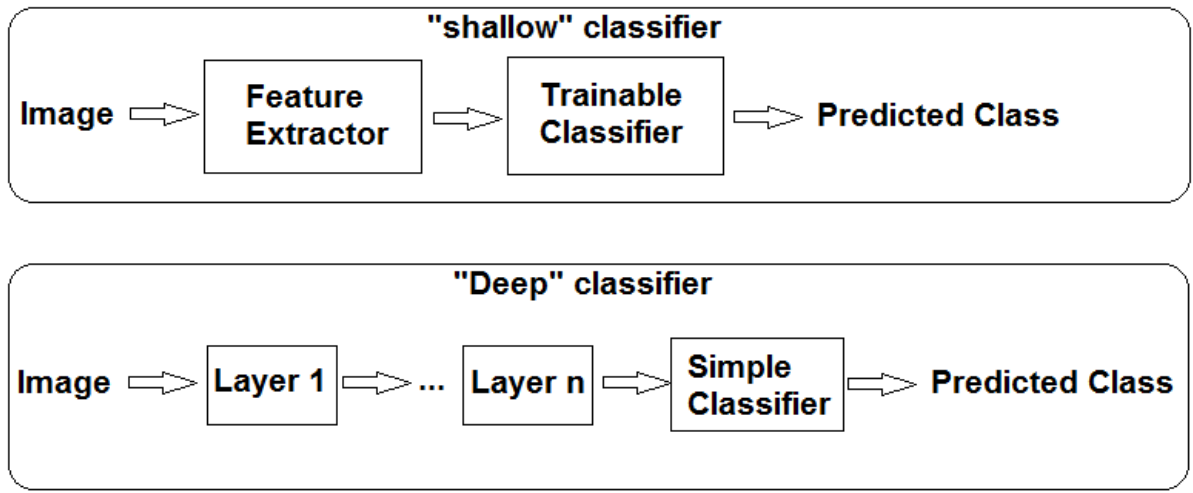

Figure 10. comparison of shallow and deep classifiers' mechanism

## 4.2 Shallow Classifiers

Shallow classifiers are classifiers that usually consist of one layer of kernel functions that will take the image features as input and compare them to previously learned patterns from the training set to finally predict a score/class. Shallow classifiers rely on an external mechanism to extract the features and the better the features the better the result. Choosing the right feature extractor will have a huge effect on the classification which can be tricky sometimes. For example the features that could perform well for face detection won't work well for scene classification. Shallow classifiers do not require a lot of hyperparameters nor do they need a lot of computational power. Shallow classifiers have decent results without the need of a large scale database to train them. One of the simplest examples is the Perceptron which is an algorithm used for binary classification. The perceptron will map an input vector $x$ to an output value $f(x)$ such that:

$$f(x) = \begin{cases} 1 & if \; w \cdot x + b > 0 \\ 0 & otherwise \end{cases} \qquad (3)$$

Where w is the value of the weights and b is the bias that can be thought of as the threshold that separates the 0s and 1s. As for the dot product $w \cdot x$ it represents the sum: $\sum_{i=0}^{N} w_i x_i$ of the entire N inputs [18].

Another good example of a shallow classifier is the Multilayer Perceptron (MLP) that is an artificial neural network that is built with multiple fully connected layers of neurons with a non-linear activation function [10]:

$$f(x) = \tanh(x) \quad or \quad f(x) = (1 + e^{-x})^{-1} \qquad (4)$$

MLP uses a supervised learning with a backpropagation technique for the training part. MLPs are a great example of so called shallow classifiers that can discriminate on non-linearly separable data but they suffer when handling high resolution images since they increase the dimensionality and reduces the predictive power [12].

## 4.3 Deep Classifiers

Deep classifiers are constructed by multi-layers of adaptive non-linear modules that encompass learnable parameters at all levels. Each layer takes as input the output of

Training Convolutional Networks with Web Images – Nizar Massouh

the previous layer and the first layer takes as input the raw image. These cascades of multi-layers will extract features while reducing the spatial size as it goes higher in the level of features' hierarchy. This means that deep classifiers can learn feature extractors independently of the task. As an example of deep classifiers we will focus on convolutional neural networks (CNNs).

## 4.4 Convolutional Neural Networks

The neural network receives the image as a vector and transforms it from one hidden layer to another (depending on the architecture). Hidden layers are made of neurons and each layer is fully connected to another through their neurons (receptive field), but neurons of the same layer are not connected. The neurons are stacked in 3 dimensions: width, height and depth. Since the neurons of adjacent layers are connected locally, they will focus on spatially local patterns. The more layers will be stacked and connected, the more the local representation will become global.

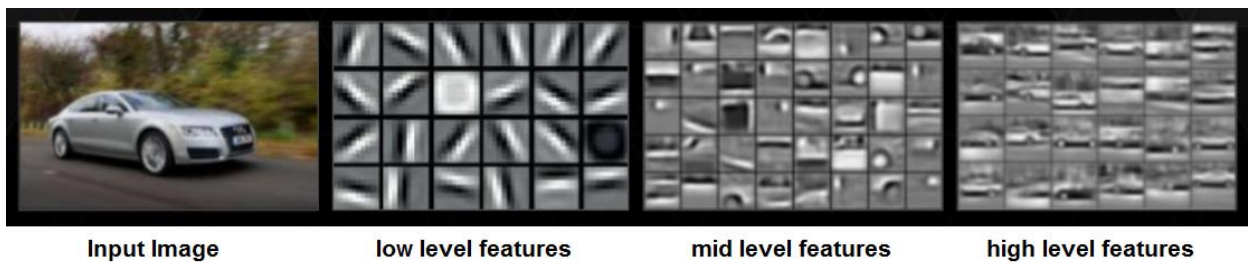

Input Image          low level features          mid level features          high level features

**Figure 11. Different levels of learned features**

Training Convolutional Networks with Web Images – Nizar Massouh

As we can see in figure 11, the low level features that were learned at the first layer show features that represent local details of the image like curves and edges. As we advance with the layers the extracted features show a more global representation of a car. This proves the feature extracting potential that a CNN has, which will be investigated with visualizations later in chapter 5 using the t-SNE algorithm [19]. T-SNE transforms Euclidean distance of the high-dimensional data-points into conditional probability $P_{j|i}$ which represents the probability of that the point $x_i$ will have $x_j$ as its neighbor. $P_{j|i}$ is high for nearby data-points and very low for faraway data-points, which can be used as similarity indicator. For the low dimensional representation of the algorithm computes a similar conditional probability $q_{j|i}$ which represents the probability of that the point $y_i$ will have $y_j$ as its neighbor. Since $y_i$ and $y_j$ model the similarity between the high dimensional points ($x_i, x_j$) then the 2 probabilities should be equal. Based on this hypothesis the algorithm next tries to minimize the mismatch between the two conditional probabilities. The last layer of the network is the output layer that should provide the class score. The architecture of CNNs is built by the stacking pattern of the different types of layers and it depends on its application. For visual recognition the architecture that was brought forward by Alex Krizhevky

（Alexnet） seems to have outstanding performances when used on a similar task as ours [11].

## 4.5 Case Study: Alexnet

The Convolutional Networks that made a breakthrough in Computer Vision was the AlexNet, developed by Alex Krizhevsky, Ilya Sutskever and Geoff Hinton. The AlexNet participated in the ImageNet ILSVRC challenge in 2012 and impressively beaten the second runner-up （top 5 error of 16% compared to runner-up with 26% error） [13]. It differs from previous networks since it was deeper, larger and included Convolutional Layers stacked on top of one another （before they used to only have one convolutional layer followed by a pooling layer）. AlexNet's architecture contains eight learned layers: five convolutional and three fully-connected distributed as follows （figure 12）:

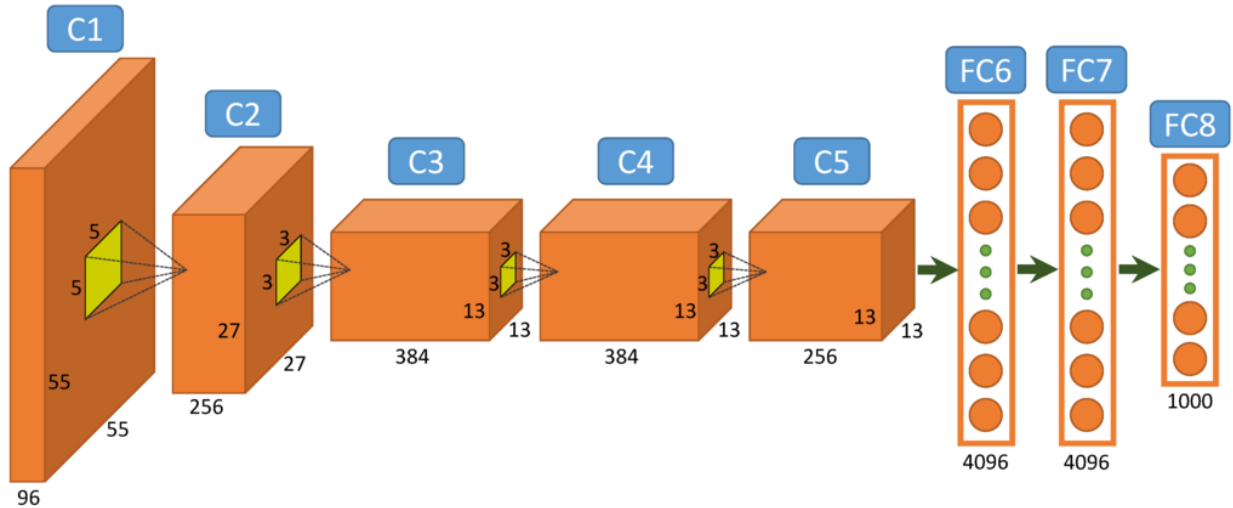

**Figure 12. Alexnet's achitecture**

**Convolutional layer:** is the main part of a CNN. This layer produces learnable kernels which have a small receptive field but goes deep into the image's volume (height x width x depth). Each kernel is convoluted through the width and height of the input's volume and computes the dot product of its entries and the input which gives the activation map of that filter. The output of the layer will be the stacked activation maps (figure 13).

For example, if our input image has size [32x32x3], if the receptive field is 8x8, then each neuron in the Convolutional Layer will get weights to a [8x8x3] region in the input volume, resulting in a total of 8*8*3 = 192 weights (and +1 bias parameter).

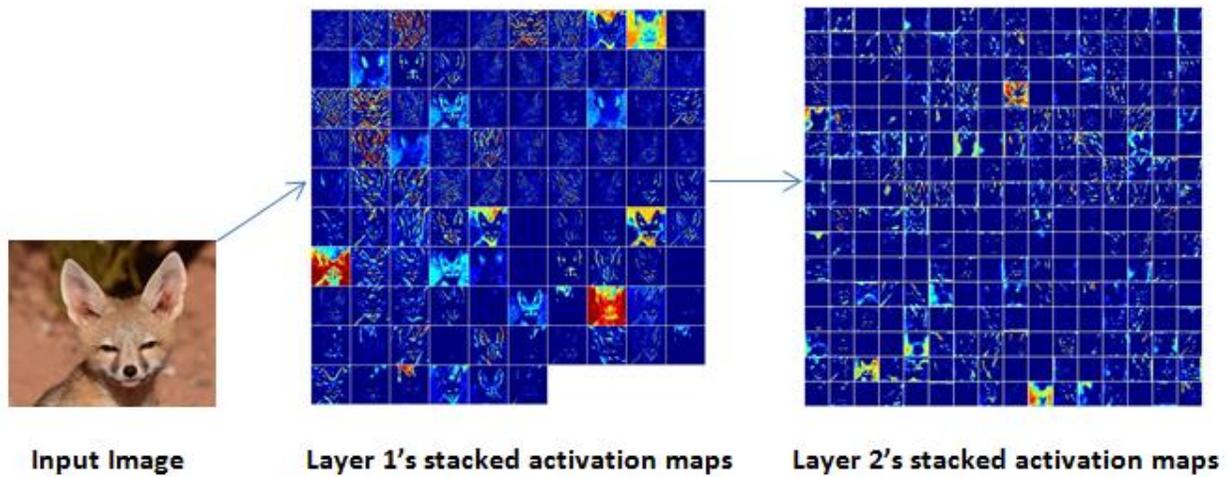

**Input Image**     **Layer 1's stacked activation maps**     **Layer 2's stacked activation maps**

Figure 13. Extracted activation maps at various layers

**Pooling layer:** is a very essential concept that is a method for down-sampling. There are several implementations and Max-pooling is among the most used. The way it works is by dividing the input image into non overlapping regions and for each it will outputs the maximum (figure 14). This means that the position of an extracted feature is relative to the region which preserves the neighbouring features' relative position. This layer will drop the spatial size to help reduce the computation by reducing parameters. Placing a pooling layer after a convolutional layer will work as a way to make the features representation more flexible.

Training Convolutional Networks with Web Images – Nizar Massouh

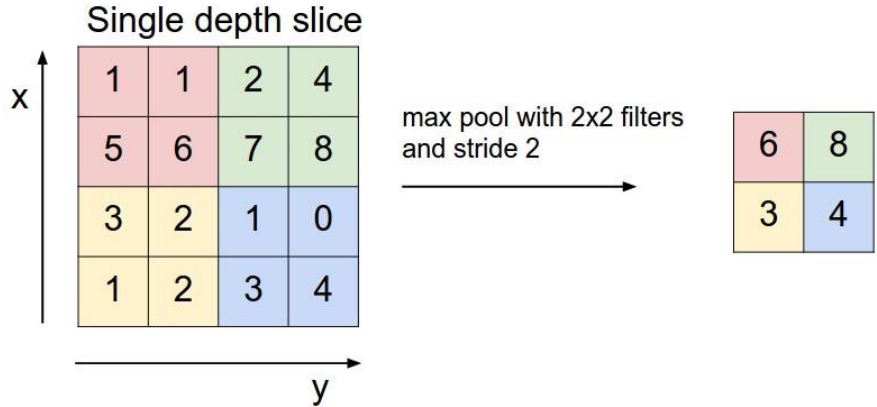

**Figure 14 Max pooling example**

**ReLU layer:** or Rectified Linear Units layer is used to increase the non−linearity of the whole network. The layer consists of neurons that apply a certain non−saturating activation function like **(4)**. ReLU layers increase the training speeds without sacrificing performance.

**Fully connected layer:** this layer comes after all the previously mentioned ones. In this layer all neurons are fully connected to all activations in the prior layers. Their activations are obtained by matrix multiplication plus a bias offset.

**Loss layer:** is the layer that works as a cost function where it provides a penalty to the ground truth and the predicted label. The most commonly used function in a loss layer is the SoftMax **(5)** [15]; which is used for predicting 1 class out of K mutually exclusive classes:

$$\boldsymbol{\sigma(z)_j} = \frac{e^{z_j}}{\sum_{k=1}^{K} e^{z_k}} \qquad \text{for } j=1,...,K \quad (5)$$

Training Convolutional Networks with Web Images − Nizar Massouh

Z in (5) is a K dimensional vector and the output of $\sigma(z)_j$ is in the range (0, 1) representing the scores of each class.

Training a CNN is a very time consuming procedure, but with the help of GPUs and tools like NVIDIA's DIGITS [20] it has come a long way. DIGITS is a deep learning tool created by NVIDIA to make the most out of their GPUs. It allows you to train, test and visualize a neural network. We have used DIGITS to conduct our experiments, by first uploading our Databases into the program and using them to train Alexnets. Our CNN is trained on our labeled data using Stochastic Gradient Decent (SGD) and the backpropagation algorithm over 30 epochs. Gradient descent (GD) is an algorithm that updates a set of parameters or weights (θ) while iterating over the data to minimize an error function. GD starts the algorithm with a random θ and repeatedly updates it using the formula:

$$\theta_j := \theta_j - \alpha \, \nabla_\theta J(\theta) \qquad j = 0 \ldots n \qquad (6)$$

In (6) α is the learning rate and the algorithm is taking steps in the direction that minimizes $J$. $\nabla_\theta J(\theta)$ is the gradient of the loss function and in (6) we are updating θ in the opposite direction of it. The learning rate α determines the step size that we take while seeking the local minimum. In GD the weights are updated after each epoch, which means after we have iterated over the whole training set. For a large scale databases GD can be costly since it only updates after each epoch which means the larger the training data the slower the algorithm gets updated. For this case SGD is very efficient since it performs a parameter update after each training example. Since SGD performs many updates with a high variance that could make the loss function to fluctuate heavily (figure 15). Having this fluctuation is handy since it might help us get to a better local minimum real fast, but at the same time it could make it more difficult to converge to the exact minimum. To help stabilize the randomness of this algorithm we decrease the learning rate every 10 epochs by 10% starting with α = 0.01.

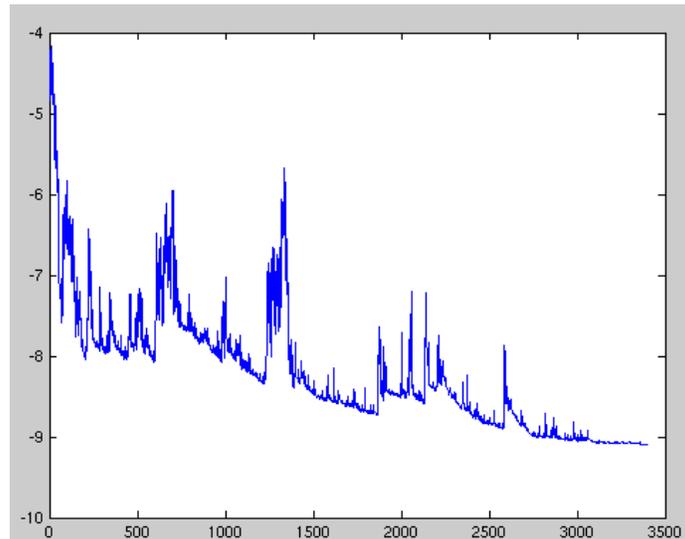

**Figure 15. SGD fluctuation [16]**

Training Convolutional Networks with Web Images – Nizar Massouh

# Chapter 5

# Experiments

In a first set of experiments we wanted to verify what performance is possible to obtain by using an AlexNet on a version of ImageNet downloaded from the web. We used different version of the noisy ImageNet, each corresponding to a different downloading strategy (for more details see section 3.5). As we derived four different versions of noisy ImageNet, we performed four different experiments. The experiments aim to evaluate the performance of our different data in terms of the accuracy on a classification task, as feature extractor for object recognition purposes and as feature extractor for domain adaptation.

## 5.1 Activation Features

As explored in [23] the layers activations of a convolutional neural network can be used as features that changes characteristics with relation to the depth of the layer. The results they showed with respect to the depth of the layers indicate that the features learn local features on the first layers and switch gradually to global feature on the last layers. We will test this theory in this section and explore our networks learning in relation to the data they were trained on. We trained a network for each of our downloaded databases, and as expected each model obtained demonstrated a different outlook when validated on the same validation set provided by [4]. As explained in section 3.5 our databases have different configurations. Specifically, we have: ***Controlled Query,*** which focuses on quality from search and image quality (High Resolution images from Google, Yahoo and Flickr) and uses a semi-supervised query expansion method that should make its images the most relevant by comparison to the other databases.

***Controlled Download*** focuses on choosing the first returned images from the search engine's results (Google and Yahoo) while using our unsupervised query expansion method. ***Random split*** uses no query expansion methods focuses on getting as many images as possible from Picsearch.com. ***Controlled split*** tries to get the images returned from the first result pages of Picsearch.com with the assumption that the relevance of the result decreases the deeper we progress with the result.

We can see in Table 2 the accuracy of the validation set. The top 1 accuracy is when the model guesses correctly the label as its top guess (highest percentage), whereas top 5 portraits the times the label is present on the first 5 guesses (does not matter what rank it is).

| - | Top-1 accuracy | Top-5 accuracy |
|---|---|---|
| Controlled Query split | **36.88%** | **58.55%** |
| Controlled download split | **34.86%** | **57.26%** |
| Controlled split | **31.06%** | **53.07%** |
| Random split 1 | **28.17%** | **50.67%** |
| Random split 2 | **28.17%** | **50.77%** |
| Random split 3 | **28.35%** | **50.88%** |

**Table 2. The top-1 and top-5 validation set accuracy of our various models**

As we can see in table 2 the accuracy of the network increases the more the data is relevant, which in a way demonstrates that our methods control the download or the query has proven to be efficient to add up to 8% of accuracy on the top-1 guess.

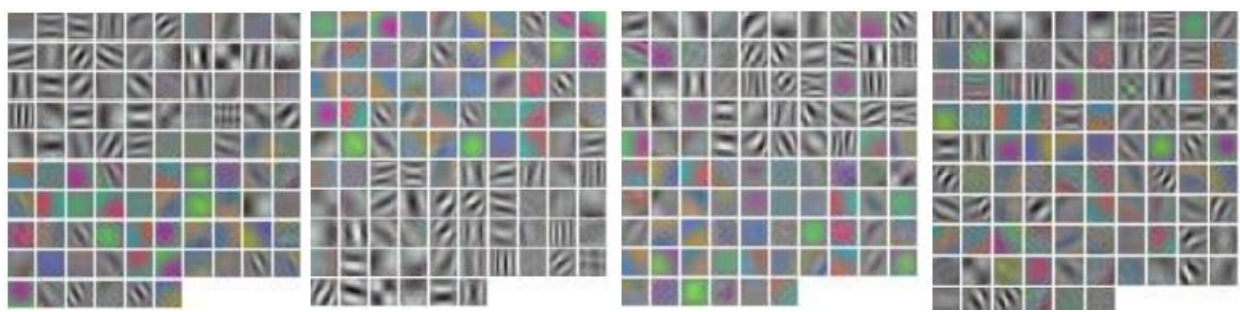

**Figure 16. The weights learned on the first convolutional layer: (from left to right) Random Split, Controlled Split, Controlled Download and Controlled Query**

Training Convolutional Networks with Web Images – Nizar Massouh

Since we have mentioned that a convolutional neural network learns the feature extractor every time it is being trained on different data, we decided to visualize the different features extracted. Using the t-SNE [19] algorithm that works well for visualizing high dimensional data by giving each data point a two dimensional position. In fact it is transforming the Euclidean distances between data points into conditional probabilities that represent similarities to finally plot them as points colored depending on their class. As we have mentioned in chapter 4, in a CNN the layers that are the closest to the output are expected to be linearly separable. In figure 16 we observe how our first layers learned low level features consisting of edges and corners which provide a very local outlook of the image. To make sure that our high level activation layers have learned good feature, taking [3] as a reference, we used the validation set of ILSVRC-2012 to create 5 super-classes consisting of a collection of 11 classes each as explained in table 3:

| Category | Class numbers | | | | | | | | | | |
|----------|-----|-----|-----|-----|-----|-----|-----|-----|-----|-----|-----|
| Dogs | 45 | 170 | 119 | 210 | 107 | 126 | 88 | 145 | 59 | 160 | 152 |
| insects | 224 | 631 | 632 | 633 | 634 | 635 | 636 | 637 | 638 | 639 | 640 |
| Birds | 388 | 389 | 390 | 391 | 392 | 393 | 394 | 395 | 396 | 397 | 398 |

| Snakes | 482 | 483 | 484 | 485 | 486 | 487 | 488 | 489 | 490 | 491 | 492 |
|--------|-----|-----|-----|-----|-----|-----|-----|-----|-----|-----|-----|
| Fruits | 746 | 318 | 319 | 229 | 320 | 321 | 322 | 323 | 324 | 325 | 326 |

**Table 3. The super-classes created to visualize the features (Check Appendix for class names)**

The number of images per category (table 3) is 550 which means we will be visualizing 2750 images of different categories. To be able to visualize the features we first have to extract them. The steps taken to produce the t-SNE plots start by feeding forward our images to be able to extract the features of the fully connected layer 6 and 7. Next the features will undergo a dimensionality reduction using Principal Component Analysis [22] before they are fed to the t-SNE algorithm that will produce the 2D representation to be plotted.

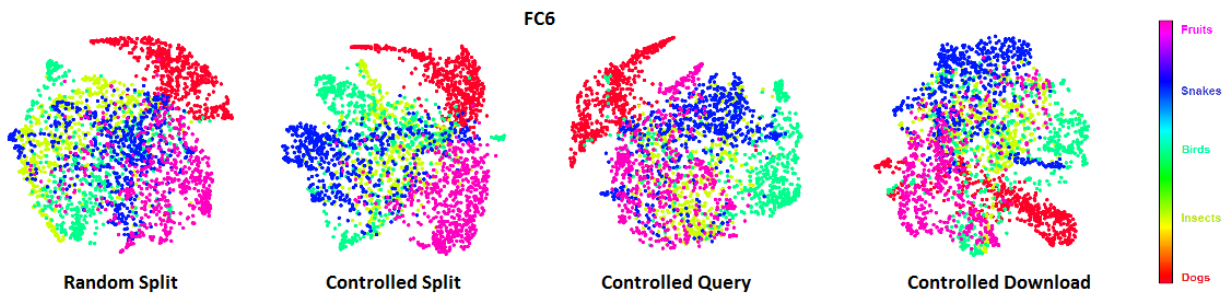

**Figure 17. This figure shows the various t-SNE feature visualizations on the ILSVRC-2012 validation set on the 6th layer (Fully Connected Layer 6).**

FC7

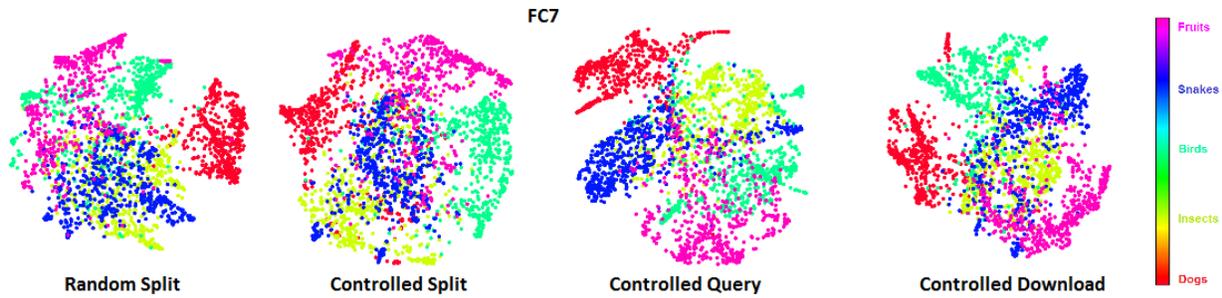

Random Split          Controlled Split          Controlled Query          Controlled Download

**Figure 18. This figure shows the various t-SNE feature visualizations on the ILSVRC-2012 validation set on the 6th layer (Fully Connected Layer 7).**

Figure 17 and figure 18 show the extracted features on our validation set super-classes using the two last fully connected layers. As we can see the two layers show very well separated features and as expected the $7^{th}$ layer has the features more spaced out which clearly captures the semantic difference in the images. Some of our super-classes seem to be clustered better, i.e. Dogs and Birds, the reason behind that is caused by the environmental similarities of the other 3 or in other words the similarity of their backgrounds. Since the Fruit category contains Mushrooms, we can see that it is partially mixed with snakes and insects which prove the background hypothesis as we can see in figure 19.

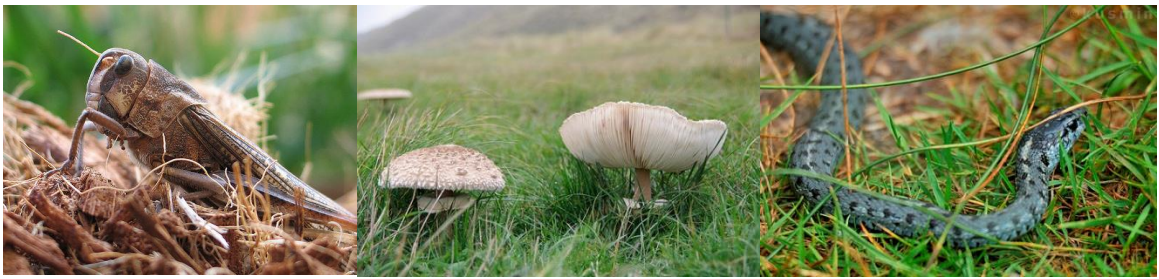

Training Convolutional Networks with Web Images - Nizar Massouh

We can see clearly that the features provided by FC6 and FC7 show good clustering even on challenging classes, which proves that they are more useful for object recognition tasks than the lower levels features (figure 16) making our results in alignment with [21].

## 5.2 Object Recognition

To be able to evaluate the full object recognition capability of our extracted features, we decided to follow [23] [21] and use Caltech-101 dataset [29]. The experiment consists of extracting the features of the $6^{th}$ and $7^{th}$ layer of the network and testing them with a simple linear classifier. The classifier used for this experiment is Support Vector Machine (SVM) [24] which when given data of two categories will try to find a hyperplane that can separate the data and to be used to classify them. Since we have a multi class classification we will be implementing the "one versus one" approach which reduces the problem to multiple binary classifications [25] [26] such that:

If we have $N$ classes then $N \times (N-1)/2$ are built and each one will train on data for two classes. Caltech 101 contains a total of 9,146 images, split between 101 distinct object categories (faces, watches, ants, pianos, etc.) and a background category. The SVM is trained on 30 samples per class (including the background category) and tested on the remaining data. In table 4 we can observe the results that are averaged over 5 data splits per category.

| Classifier | Caffe Model | Layer | Mean Accuracy | Decaf 2014 [3] |
|---|---|---|---|---|
| SVM | Controlled Download | Fc7 | 85.79±0.4 | 83.24±1.2 |
| | | Fc6 | 87.79±0.4 | 84.77±1.2 |
| | Controlled Split | Fc7 | 85.59±0.4 | 83.24±1.2 |
| | | Fc6 | 88% | 84.77±1.2 |
| | Random Split 3 | FC7 | 84.59±0.4 | 83.24±1.2 |
| | | FC6 | 86.79±0.4 | 84.77±1.2 |
| | Controlled Query | FC7 | **86%** | 83.24±1.2 |
| | | FC6 | **88.2±0.4** | 84.77±1.2 |

Table 4. Average accuracy per class on Caltech-101with 30 training per class using the features of FC6 and FC7 compared with [21]

As we can see in table 4 our network outperformed [21] on both FC6 and FC7 by about 3%. This could only show how much effect the data has on the features. We

Training Convolutional Networks with Web Images – Nizar Massouh

can observe that even our Random split data that contains a lot of noise has performed better than Imagenet. The improvement in performance from our Random split to our Controlled Query split, which is about 2%, comes from the fact that the network were trained on an object recognition oriented task so the relevance of the data played a good role in filtering out the bad features.

## 5.3 Domain Adaptation

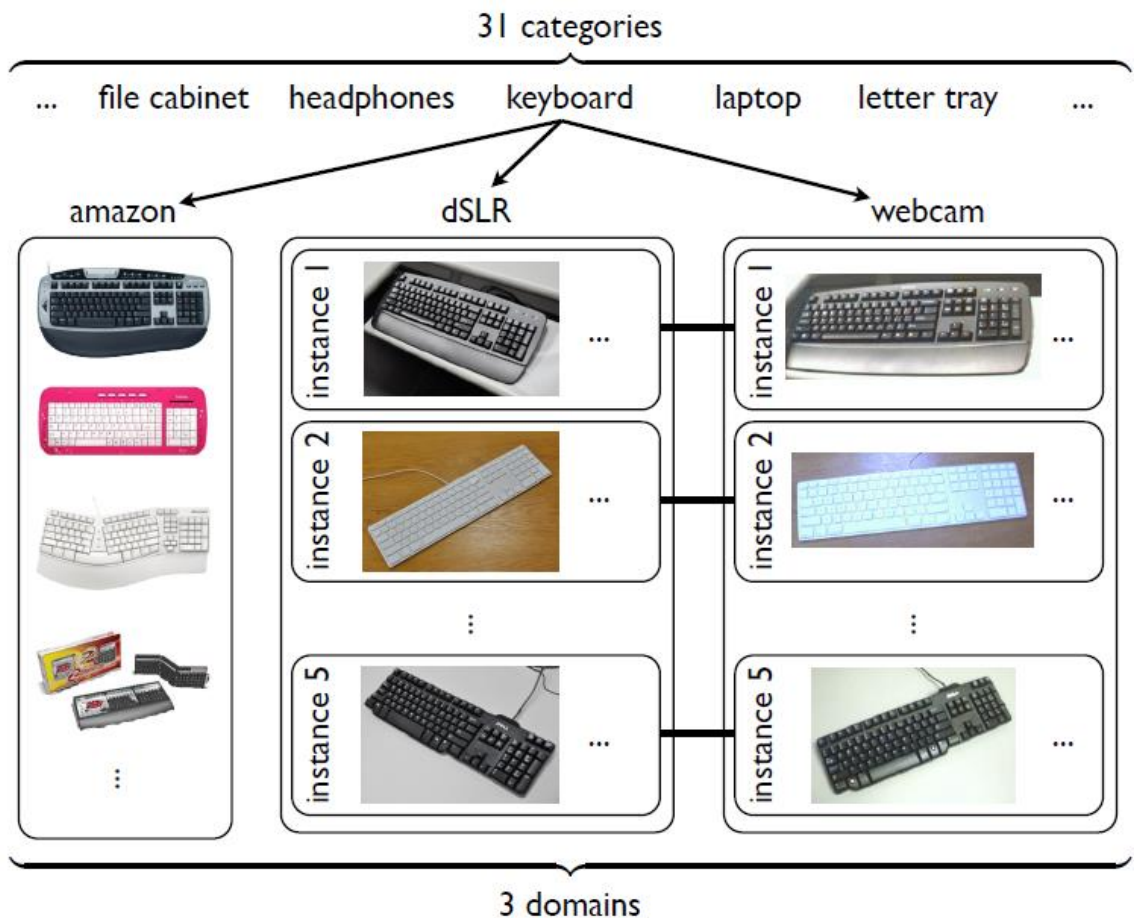

**Figure 20. Examples from the different domains in the Office database [9]**

Training Convolutional Networks with Web Images – Nizar Massouh

Another important aspect to investigate has to be how well do our features generalize when switching domains? Domain adaptation is when we have a module trained on data for a source and then tested on a target domain different from the source. Using as benchmark [21] we replicated the experiment from [27]. Our experiment will be conducted on the Office dataset that contains 3 different domains: each containing the same 31 categories: Amazon, Webcam and DSLR (figure 20). Amazon consists of images collected from the web (www.amazon.com) which is by the closest to our data. Webcam's images are recorded using a webcam which means they are noisy because of their low resolution. The last domain is DSLR, which is filled with images captured with a digital SLR camera which have a very high resolution (4288x2848). Just like [21] we will be examining the domain shifts: Amazon→Webcam and Dslr→Webcam. Using features extracted from our networks we will train an SVM on the source domain and test it on the target domain. SVM will be trained in three different ways: (S) source only, (T) target only and (ST) source and target data. Following the experimental setup of [9] the experiment was semi-supervised where 3 labels per target class were given and 20 labels per source for Amazon or 8 if the source is Dslr.

Table 5 presents the multi-class accuracy averaged across 5 train/test splits over the 2 domain shifts `Amazon→Webcam` and `Dslr→Webcam`. As we can see the result is very interesting, we have outperformed [21] every time when the source domain was Amazon and our result seems to be more consistent when switching domains. For example they had about 35% difference when switching sources for `SVM(S)` whereas we had 0% for FC6 and almost 8% for FC7. The result makes sense since our data was collected from the web and so is Amazon which means that we should be sharing some similar characteristics on the feature level. Having consistent results, while switching domains, means that our networks were able to learn generalized features with our data better than Imagenet. Having used a mixture of search engines (Google, yahoo, flickr and picsearch) help our networks not to grasp a bias concept of an appearance. While google tends to provide "cleaner" images with controlled settings, Flickr provides more difficult images with cluttered backgrounds or multiple objects present. Other work that involved multiple search engines [28] noted the differences in difficulties of images and was able to use that to their advantage as well.

| Classifier | Caffe Model | Layer | Amazon→Webcam | Dslr→Webcam | Classifier | Amazon→Webcam | Dslr→Webcam | Classifier | Amazon→Webcam | Dslr→Webcam |
|---|---|---|---|---|---|---|---|---|---|---|
| **SVM (S)** | **Random Split 1** | **Fc6** | 90% | 90% | **SVM (T)** | 99% | 99% | **SVM (ST)** | **82±1.4** | 81.59±0.48 |
| | | **Fc7** | 90% | 82.4±1.74 | | 73.2±0.4 | 73.2±0.4 | | **90%** | 90% |
| | **Random Split 2** | **Fc6** | 90% | 90% | | 98.79±0.4 | 98.79±0.4 | | 81.4±0.48 | 81.79±0.74 |
| | | **Fc7** | 90% | 82.79±0.74 | | **74±0.63** | **74±0.63** | | 90% | 90% |
| | **Random Split 3** | **Fc6** | 90% | 90% | | 98.79±0.4 | 98.79±0.4 | | 81.79±1.16 | 81±0.63 |
| | | **Fc7** | 90% | 83.2±0.97 | | 73% | 73% | | 90% | 90% |
| | **Controlled Split** | **Fc6** | 90% | 90% | | **99.2±0.4** | **99.2±0.4** | | 81.59±1.01 | 81.4±1.2 |
| | | **Fc7** | 90% | 81.79±0.74 | | 73.2±0.4 | 73.2±0.4 | | 90% | 90% |
| | **Controlled Download** | **Fc6** | 90% | 90% | | 98.59±0.48 | 98.59±0.48 | | 81.59±0.8 | 82±0.63 |
| | | **Fc7** | 90% | 81±0.63 | | 73% | 73% | | 90% | 90% |
| | **Controlled Query** | **Fc6** | **90%** | 90% | | 98.79±0.4 | 98.79±0.4 | | 81.59±1.01 | 81.59±0.48 |
| | | **Fc7** | **90%** | 81±1.67 | | 73.2±0.4 | 73.2±0.4 | | 90% | 90% |
| | **Decaf 2014 [3]** | **Fc6** | 52.2±1.7 | **91.48±1.5** | | 78.2±2.6 | 78.2±2.6 | | 80.66±2.3 | **94.79±1.2** |
| | | **Fc7** | 53.9±2.2 | **89.15±1.7** | | 79.1±2.1 | 79.1±2.1 | | 79.12±2.1 | **92.96±2.0** |

**Table 5. Presents the multi-class accuracy averaged across 5 train/test splits over the 2 domain shifts Amazon→Webcam and Dslr→Webcam.**

Training Convolutional Networks with Web Images − Nizar Massouh

## 5.4 Discussion

In this chapter we were able to demonstrate the many traits of a trained CNN. We were able to see that having noisy data doesn't help the accuracy of a classification task (table 2) but plays a big role in helping the network learns a real general concept of the object. We saw how just by using a simple query expansion / relevance method we were able to enhance the accuracy by a stunning 8% from the Random Split to the Controlled Query split. Through the t-SNE visualization we saw how our networks learned linearly separable features that present a good global outlook of the data on the last layers (FC6 and FC7) whereas the first layer learned very locally oriented features. Using the trained networks as feature extractors has proved to be very efficient for the task of object recognition and managed to outperform the state of the art [21]. Having strong features that can represent a concept globally is exactly what is needed to recognize objects and that was evident in our results (table 4).

Another important task that can benefit from strong features is the Domain Adaptation problem. We have tested our data on the benchmark of Domain Adaptation and we were able to compete and outperform the features learned from Imagenet (table 5).

Training Convolutional Networks with Web Images – Nizar Massouh

What was more interesting was that our features managed to give consistent results that were not affected by the domain switch.

# Chapter 6

# Conclusion

This thesis explored up to which point it is possible to substitute manually annotated images with those downloaded from the web as a result of a search guided by a given query, where such downloaded images are expected to bring a considerable amount of noise. To this end, we attempted to replicate ImageNet, and we proposed four different search strategies, corresponding to four different Webly derived versions of ImageNet. We trained on each of them a CNN architecture that proved successful on ImageNet, and analysed its performance in terms of the accuracy obtained on the specific classification task, as feature extractor for the object recognition problem, and as feature extractor for the domain adaptation problem. As we saw the result was impressive since we were able to produce very strong features that are linearly separable and are robust since they were able to generalize from one domain to another. We were able to demonstrate that the use human annotators to create a database, is not always a critical factor to learn the best features. In this thesis we were able to provide a mechanism to build any database directly from web images. We proposed a query expansion method that can be used for data augmentation purposes. We showed that for the task of object recognition and domain adaptation we can compete and outperform the state of the art without relying on human annotators or expensive databases. We hope that through this work we can take a step toward a more "Plug-and-Play" approach to object recognition, where all you need is an internet connection and a concept to learn. There are still many things to investigate like the effect our query expansion method on the relevance; it would be interesting to see how much we can improve the accuracy if we expand our method to be more selective. Another task that can give a deeper understanding of how well the features can generalize, we can try more domain shifts to compare the result. As for using our features from object recognition there's still so many ways to explore the quality of our features with other databases that could be more challenging than Caltech-101.

Training Convolutional Networks with Web Images – Nizar Massouh

Training Convolutional Networks with Web Images — Nizar Massouh

Training Convolutional Networks with Web Images – Nizar Massouh

Training Convolutional Networks with Web Images – Nizar Massouh